\documentclass[conference]{IEEEtran}
\IEEEoverridecommandlockouts

\usepackage{cite}
\usepackage{amsmath,amssymb,amsfonts}
\usepackage{algorithmic}
\usepackage{graphicx}
\usepackage{textcomp}
\usepackage{xcolor}
\def\BibTeX{{\rm B\kern-.05em{\sc i\kern-.025em b}\kern-.08em
    T\kern-.1667em\lower.7ex\hbox{E}\kern-.125emX}}

\makeatletter
\newcommand\subsubsubsection{\@startsection{paragraph}{4}{\z@}{-1ex\@plus -0.5ex \@minus -.25ex}{1.ex \@plus .25ex}{\normalfont\normalsize\bfseries}}
\newcommand\subsubsubsubsection{\@startsection{subparagraph}{5}{\z@}{-1ex\@plus -0.5ex \@minus -.25ex}{1.ex \@plus .25ex}{\normalfont\normalsize\bfseries}}

\makeatother  

\def\BibTeX{{\rm B\kern-.05em{\sc i\kern-.025em b}\kern-.08em
    T\kern-.1667em\lower.7ex\hbox{E}\kern-.125emX}}

    \def\argmin{\operatornamewithlimits{arg\,min}}
    \usepackage{longtable}
\usepackage{hyperref}
\usepackage{caption}

\usepackage{float}

\begin{document}

\title{ \Large A Real-Time Generalized Nash Equilibrium Framework for Interaction-Aware Autonomous Driving in Mixed Traffic\\
}







\author{

\IEEEauthorblockN{\textbf{Nouhed Naidja}}
\IEEEauthorblockA{
\textit{Institut VEDECOM, Versailles, France}\\
nihed.naidja@vedecom.fr
}
\and

\IEEEauthorblockN{Mohamed-Cherif Rahal}
\IEEEauthorblockA{
\textit{Institut VEDECOM, Versailles, France}\\
mohamed.rahal@vedecom.fr
}
\and

\IEEEauthorblockN{Steve Pechberti}
\IEEEauthorblockA{
\textit{Institut VEDECOM, Versailles, France}\\
steve.pechberti@vedecom.fr
}

\and
\IEEEauthorblockN{Stéphane Font}
\IEEEauthorblockA{
\textit{Laboratoire des Signaux et Systèmes (L2S)}\\
stephane.font@centralesupelec.fr
}

\and\and\and
\IEEEauthorblockN{Guillaume Sandou}
\IEEEauthorblockA{
\textit{Laboratoire des Signaux et Systèmes (L2S)}\\
guillaume.sandou@centralesupelec.fr
}
\and

\IEEEauthorblockN{Marc Revilloud}
\IEEEauthorblockA{
\textit{Dotflow}\\
marc.revilloud@dotflow.fr
}

}

\maketitle
\begin{abstract}
Safe and efficient navigation in mixed-traffic environments remains a critical challenge for Autonomous Vehicles (AVs), primarily due to the complex interdependence between the AV's decisions and the unpredictable reactions of human drivers. This paper introduces a comprehensive decision-making framework that formulates the driving interaction as a Generalized Nash Equilibrium Problem (GNEP). Unlike decoupled optimization approaches, this framework explicitly models shared safety and geometric constraints, ensuring that the feasibility of the AV’s strategy is dynamically linked to the opponent’s actions. To solve this non-convex problem in real-time, we propose a dedicated solver based on Particle Swarm Optimization (PSO). The complete architecture was validated on a test track using a real autonomous Renault Zoé interacting with a human driver. Experimental results demonstrate the system's ability to handle critical scenarios by generating comfortable, human-like trajectories. Benchmarks confirm the solver’s operational feasibility, achieving convergence in under 50 ms.
\end{abstract}
\section{Introduction}
As the transition towards autonomous mobility progresses, the interaction between autonomous vehicles (AVs) and human drivers will create unique traffic dynamics. In this context, the paper in \cite{mahajan2021intention} lists a number of human drivers' expectations regarding the behavior of autonomous vehicles. They expect autonomous vehicles to: (i) be aware of their intentions, (ii) act consistently with respect to their expectations, and also (iii) exhibit a behavior recognizable and understandable by human drivers. 
Failure to meet these expectations could lead to mistrust, reticence, and even a rejection of cohabitation with autonomous vehicles.
In such a context, it is no longer sufficient for autonomous vehicles to merely adhere to rules or optimize predefined objectives independently.
They need to be able to understand, predict, and adapt to the behavior of others.
These observations highlight the need for a decision-making framework that will enable autonomous vehicles to choose actions that are not only safe and efficient but also interaction-aware. 
Game-theoretic-based approaches have particularly drawn researchers' interest, aiming to formalize and analyze the interdependencies that may arise in transportation systems \cite{qin2024game}.
In line with this research direction, this paper proposes a decision-making framework, grounded in game theory, that shifts the paradigm from egocentric optimization to multi-agent interactive planning. By explicitly modeling the coupled nature of traffic, we aim to bridge the gap between algorithmic safety and human-like negotiation.

\section{Background}
\label{sec:Background}
Driving is inherently game-theoretic: each driver's decisions depend on predictions 
of others' reactions, creating a complex interdependent decision-making environment 
\cite{qin2024game}. Early formulations model such interactions as 
two-player zero-sum games \cite{zhang2021safe,bacharach1989zero}, combining proactive 
and reactive strategies for robust decision-making. However, traffic scenarios such 
as merging or intersection crossing do not fit the zero-sum framework, as mutual 
benefit is possible. Non-zero-sum dynamic games \cite{8643396} better capture these 
collaborative aspects by allowing simultaneous gains for all players.
Within game theory, the Nash equilibrium defines a strategy profile where no player 
can unilaterally improve their payoff given fixed strategies of others 
\cite{nash1950equilibrium}. Yet, this classical formulation assumes independent 
strategy sets, which is rarely satisfied in traffic due to shared safety constraints 
and road geometry \cite{lei2022stochastic}. To address these dependencies, the Nash 
equilibrium is extended to Generalized Nash Equilibrium Problems (GNEPs), where both 
objective functions and feasible strategy sets depend on the opponents' strategies 
\cite{facchinei2010generalized}.

The remainder of this paper is organized as follows: Section~\ref{sec:Methods} 
details the proposed decision-making model, Section~\ref{sec:Meta-Heuristics} 
develops the PSO-based solver, and Section~\ref{sec:XP} assesses its performance 
through simulation experiments.
\section{Methods}
\label{sec:Methods}

\subsection{Multi-Agent Interaction as a Game}
We frame the autonomous vehicle’s decision problem as a two-player, non-zero-sum game. The Ego vehicle, denoted \(v\), chooses a strategy \(s_v \in \mathbb{R}^{n}\) within a feasible strategy set \( \mathcal{S}_{v}\).
The opponent's strategy is denoted \( s_o \in \mathbb{R}^{m}\).
Where $n$ and $m$ are the decision variables, characterizing the vehicle's motion ( control point of trajectory, velocity profile parameters). 
Considering this mutual interdependence, we define a joint strategy space \( \mathcal{S}_{\text{joint}} = S_v \times S_o \subseteq \mathbb{R}^{n+m} \) that encompasses all possible pairs of feasible strategies \( (s_v, s_o) \), and provides the domain in player's interactions can be jointly analyzed.
Within this strategic space, each player  \( i \in  \left\{  v,o \right\} \) incurs a cost \( Q_i(s_v, s_o): \mathbb{R}^{n+m} \to \mathbb{R}\) that it aims to minimize over its feasible strategy set, given the strategy of the other player.

Within this strategic space, each player \(i \in \{v,o\}\) incurs a cost \(Q_i(s_v,s_o)\) that it seeks to minimize.  
In our formulation, the structure of \(Q_i\) is derived directly from our previous work \cite{naidja2025toward}, where we employ a multi-criteria approach to evaluate the quality of a generated trajectory.


Specifically, each cost function combines three components:
\begin{itemize}
    \item \textbf{Safety:} geometric interaction metrics between adaptive elliptical 
    safety zones, capturing minimum-distance and overlap severity.
    \item \textbf{Comfort:} longitudinal and lateral jerk indicators evaluated along 
    the trajectory.
    \item \textbf{Efficiency:} global travel time to complete the maneuver.
\end{itemize}
These components are aggregated via a weighted sum, complemented by jerk constraint 
penalty terms. The functions \(Q_v\) and \(Q_o\) share this structure, each evaluated 
from the respective agent's perspective given the opponent's predicted strategy.

Then, we assume that each player is motivated to choose the best response available to optimize its own utility \cite{askari2019behavioral}.
Let the joint best response operator \( {BR}\) define a mapping which assigns to each pair of strategies the corresponding pair of optimal responses.
\begin{equation}
\begin{aligned}
BR : S_v \times S_o &\rightarrow S_v \times S_o \\
(s_v,s_o) &\mapsto \left(BR_v(s_o),BR_o(s_v)\right)
\end{aligned}
\label{eq:Best_resp}
\end{equation}

Considering the best response concept, \( {BR}_v\)  and \( {BR}_o\) correspond to the individual optimization problems solved by each player. Specifically, given the opponent's choice, the best responses are:   
\begin{equation}
BR_v(s_o) = \argmin_{s_v \in S_v} Q_v(s_v, s_o)
\label{eq:BR_V}
\end{equation}
\begin{equation}
BR_o(s_v) = \argmin_{s_o \in S_o} Q_o(s_v, s_o).
\label{eq:BR_o}
\end{equation}
Here \( {BR}_v\) defines a mapping from \(S_o\) to \(S_v\), assigning to each strategy of player $o$ the optimal response of player $v$. Likewise, \( {BR}_o\) defines a mapping from \(S_v\) to \(S_o\).
The joint best response operator in Equation \eqref{eq:Best_resp} aggregates the two individual best responses Equations \eqref{eq:BR_V} and \eqref{eq:BR_o}. 
It assigns to every pair \( (s_v, s_o)\) the corresponding pair of mutually optimal responses.

\subsection{Shared Constraints and Feasible Strategy Sets}
Unlike standard optimization problems, where decision variables are independent, in our problem, the strategies of the ego and opponent vehicles are coupled through safety, kinematic, and road geometry constraints.This means that the feasibility of a strategy of an agent depends directly on the choice of the other one.
This coupling is expressed through a set of shared inequality constraints of the form: 
\begin{equation}
    h(s_v, s_o) \le 0,
    \label{eq:shared_constraint}
\end{equation}

where \(h : \mathbb{R}^n \times \mathbb{R}^m \rightarrow \mathbb{R}^k\) denotes a vector-valued function capturing all constraint components (collision avoidance, road boundaries, dynamic limits, etc.).  
Such constraints directly involve both players’ strategy variables, and therefore make each player's feasible strategy set explicitly dependent on the opponent’s decision.
Formally, the feasible sets become:
\begin{align}
    \mathcal{S}_v(s_o) &= \{\, s_v \in \mathcal{S}_v \mid h(s_v, s_o) \le 0 \,\}, \\
    \mathcal{S}_o(s_v) &= \{\, s_o \in \mathcal{S}_o \mid h(s_v, s_o) \le 0 \,\}.
\end{align}
The sets \(\mathcal{S}_v(s_o)\) and \(\mathcal{S}_o(s_v)\) arise from the shared constraints \eqref{eq:shared_constraint} and illustrate the essence of a Generalized Nash Equilibrium Problem: each player optimizes over a feasible region that itself depends on the opponent’s strategy.

Considering that each player's strategy set \(S_i\) may depend on the strategies of other players \(s_{-i}\), a strategy profile \(s^* = (s_1^*, s_2^*, \dots, s_n^*)\) is a generalized Nash Equilibrium if, for every player \( i \in  \left\{  v,o \right\} \):
\begin{equation}
 \forall s_i \in S_i(s_{-i}^*), \quad s_i^* \in S_i(s_{-i}^*), \quad Q_i(s_i^*, s_{-i}^*) \geq Q_i(s_i, s_{-i}^*)
 \label{eq:Cost_fun}
\end{equation}

A generalized Nash equilibrium is then characterized by a pair $(s_v^*, s_o^*) \in S_v \times S_o$ of this mapping satisfying:
\begin{equation}
(s_v^*, s_o^*) = BR (s_v^*, s_o^*)
 \label{eq:Best_resp_strat}
\end{equation}
This definition ensures that each player's optimal strategy is not only the best response to the strategies of others but also respects the shared constraints that link the players' strategy. 
\subsection{Characterizing the Generalized Nash Equilibrium
}
\subsubsection{\textbf{Deviation Analysis and Equilibrium Consistency}}
Now, we want to evaluate whether an agent could unilaterally improve its outcome by deviating. 
To capture this dynamic, we assess how an arbitrary strategy pair $(s_v, s_o) $ relates to the equilibrium condition $(s_v^*, s_o^*)$. If $(s_v, s_o) $ coincide with $(s_v^*, s_o^*)$, then agents' unilateral deviation is not beneficial. To capture this dynamic, we define a deviation measure \( \Delta(s_v, s_o)\), one for each player,  to measure the largest possible payoff improvement that a player could obtain by switching from a strategy \(\boldsymbol{{s}_i} \) to an alternative one  \(  \boldsymbol { \bar{{s}_{v}}} \), while the other player keeps its strategy fixed. 
The deviation measures are defined as follows:  
\begin{equation}
    \Delta_v(s_v, s_o) = \sup_{\bar{s}_v \in \mathcal{S}_{v}} \left( Q_v(s_v, s_o) - Q_v(\bar{s}_v, s_o) \right)
    \label{eq:dev_v}
\end{equation}
\begin{equation}
    \Delta_o(s_v, s_o) = \sup_{\bar{s}_o \in \mathcal{S}_{o}} \left( Q_o(s_v, s_o) - Q_o(s_v, \bar{s}_o) \right)
    \label{eq:dev_o}
\end{equation}

Equation \eqref{eq:dev_v} represents the largest possible payoff improvement for player \(\textbf{v}\) if it switches from its current strategy \(\boldsymbol{{s}_v} \) to another strategy \(  \boldsymbol { \bar{{s}_{v}}} \), while the player \(\textbf{o}\) keeps its strategy fixed at \(\boldsymbol{{s}_{o}} \).
Similarly, the measure \eqref{eq:dev_o} represents the payoff improvement for player \( \textbf{o}\) when switching from  \( \boldsymbol{s_{o}}\) to \(  \boldsymbol { \bar{{s}_{o}}} \), while keeping \( \boldsymbol {s_{v}}\) unchanged. 
From the player \( \textbf{v}\)'s perspective, if the difference \( \left( Q_v(s_v, s_o) - Q_v(\bar{s}_v, s_o) \right) \) is positive, it indicates that it could benefit (minimize its cost) from deviating to an alternative strategy \(\boldsymbol{{s}_{v}} \).
Conversely, a negative difference indicates that the current strategy \( s_v \) is already preferable to \( \boldsymbol {\bar{s}_v }\), and thus there is no incentive to deviate.
The same applies to the player \(\textbf{o}\) in Equation \eqref{eq:dev_o}.
\\
\subsubsection{\textbf {Deviation Measures under Shared Feasibility}}

In the generalized Nash setting, the feasible strategy sets depend on the opponent’s decision through the shared constraints introduced earlier.  
To remain consistent with this structure, we restrict unilateral deviations to strategies that satisfy the coupled feasibility conditions.  Accordingly, the deviation measures are extended as follows: 

\begin{equation}
\Delta_v(s_v, s_o) =
\begin{cases}
\displaystyle 
\sup_{\bar{s}_v \in \mathcal{S}_v(s_o)}
\Bigl(
Q_v(s_v, s_o)
- Q_v(\bar{s}_v, s_o)
\Bigr),
\\[0.8ex]
\hspace{2.5em}
\text{if } (s_v, s_o) \in \mathcal{S}_{\text{joint}},
\\[1ex]
+\infty,
\quad \text{otherwise}.
\end{cases}
\label{eq:dev_v_gne}
\end{equation}
With an analogous definition for $\Delta_o(s_v, s_o)$.
This compact formulation enforces that only deviations respecting the shared feasibility constraints are admissible, while infeasible configurations are penalized.

When considered separately, the deviation measures \(\Delta_v(s_v, s_o)\) and \(\Delta_o(s_v, s_o)\) only capture each player's unilateral incentive to deviate, and provide information on a strategy profile concerning a single agent at a time. Thus, they don't provide information regarding the joint dynamics under which the two players simultaneously converge towards equilibrium. 
However, a Nash equilibrium is defined precisely as a configuration in which neither player has an incentive to deviate simultaneously. 
Consequently, combining both perspectives into a unified metric is necessary.
In light of these considerations, the next section introduces an approach that simultaneously searches for a pair of strategies $(s_v^*, s_o^*)$ satisfying both players' optimality conditions and the joint feasibility constraints.

\subsection{Characterizing Nash Equilibria via a Joint Criterion}
We propose formulating the autonomous vehicle’s decision-making task as a joint optimization problem.
To this end, we formulate the generalized Nash equilibrium such that the ego player, the vehicle \(v\), conceptually solves the optimization problems not only for itself but also for its opponent, the vehicle \( o\). 
In this framework, we define a joint criterion \(\mathit{J} ({s}_{v},{s}_{o})\), expressed as the aggregation f two deviation measures \( \Delta_v\) and \( \Delta_o\), as follows: 
\begin{equation}
\begin{aligned}
\mathit{J} ({s}_{v},{s}_{o}) &= \Delta_v ({s}_{v},{s}_{o})+\Delta_o({s}_{v},{s}_{o}) \\
\end{aligned}
\label{equ: Opti_All_1}
\end{equation}

In the definition \eqref{equ: Opti_All_1}, each deviation measure \( \Delta_v(s_v, s_o) \) and \( \Delta_o(s_v, s_o) \) is defined using a supremum over all possible strategies, including the current strategy \( s_v \) (respectively \( s_o \)). This guarantees that the difference \( Q_v(s_v, s_o) - Q_v(s_v, s_o) = 0 \) is taken into account, thus ensuring that the supremum is non-negative.
Thus, we have: 
\begin{equation}
\Delta_v(s_v, s_o) \geq 0 \quad \text{and} \quad \Delta_o(s_v, s_o) \geq 0
\end{equation}
\begin{equation}
J(s_v, s_o) = \Delta_v(s_v, s_o) + \Delta_o(s_v, s_o) \geq 0
\end{equation}

To make this property explicit, we introduce the maximum operator \( \max(\cdot, 0) \) as follows:
\begin{equation}
J(s_v, s_o) = \max\left( \Delta_v(s_v, s_o), 0 \right) + \max\left( \Delta_o(s_v, s_o), 0 \right)
\label{eq:potential_function_add}
\end{equation}
The criterion $J(s_v, s_o)$ Eq. \eqref{eq:potential_function_add} represents the sum of the maximum unilateral improvements available to both players; it measures the improvement of deviating from the actual strategy profile.
Consequently, when $J(s_v, s_o)$ is close to zero, the opportunity for improvement is minimal for both players. In this sense, the value of $J(s_v, s_o)$ provides an intuitive notion of the distance between the current strategy profile and a Nash equilibrium. 

Consequently, the search for an equilibrium can be framed as an optimization problem over the joint strategy space as follows: 
\begin{equation}
 \min_{\bar{s}_v \in S_v, \ \bar{s}_v \neq s_v \atop \bar{s}_o \in S_o, \ \bar{s}_o \neq s_o } J(s_v, s_o)
 \label{eq:OptiProb}
\end{equation}

The optimization problem Eq. \eqref{eq:OptiProb} may be computationally intensive, since the criterion involves an evaluation
over two nested supremum operations, leading to a minimax formulation, which can be very costly. 
To accommodate the criterion for numerical evaluation, we introduce a sequence of simplification and approximation.

We first exclude the trivial comparison \( \bar{s}_v = s_v \) and \( \bar{s}_o = s_o \), therefore reducing redundancy in the deviation measures. We redefine the criterion as:
\begin{equation}
    \begin{aligned}
        J(s_v, s_o) = &\underbrace{\max \left( \sup_{\bar{s}_v \in S_v, \, \bar{s}_v \neq s_v} \{ Q_v(s_v, s_o) - Q_v(\bar{s}_v, s_o) \}, 0 \right)}_{\Delta_v(s_v, s_o)} \\
        &+ \underbrace{\max \left( \sup_{\bar{s}_o \in S_o, \, \bar{s}_o \neq s_o} \{ Q_o(s_v, s_o) - Q_o(s_v, \bar{s}_o) \}, 0 \right)}_{\Delta_o(s_v, s_o)}
    \end{aligned}
    \label{eq:potential_function_new}
\end{equation}

Since the supremum Eq. \eqref{eq:potential_function_new} is evaluated over continuous strategy spaces, it may be computationally demanding. This is particularly problematic in optimization contexts, where the evaluation of \( J \) may need to be repeated during the search for an optimal solution.
To tackle this problem, we suggest approximating the criteria by considering a finite set of sampled strategies. Formally, we replace the supremum operators with a maximization over sampled subsets \( \hat{S}_v \subset S_v \) and \( \hat{S}_o \subset S_o \).
This yields the following discrete deviation measures: 
\begin{equation}
\hat{\Delta}_v(s_v, s_o) = \max\left( \max_{\bar{s}_v \in \hat{S}_v, \, \bar{s}_v \neq s_v} \left( Q_v(s_v, s_o) - Q_v(\bar{s}_v, s_o) \right), 0 \right) 
\label{eq:Delta_E_Appro}
\end{equation}
\begin{equation}
\hat{\Delta}_o(s_v, s_o) = \max\left( \max_{\bar{s}_o \in \hat{S}_o, \, \bar{s}_o \neq s_o} \left( Q_o(s_v, s_o) - Q_o(s_v, \bar{s}_o) \right), 0 \right)
\label{eq:Delta_O_Appro}
\end{equation}
From these, the approximated joint criterion is given by:
\begin{equation}
\hat{J}(s_v, s_o) = \hat{\Delta}_v(s_v, s_o) + \hat{\Delta}_o(s_v, s_o)
\label{eq:JhateAppro}
\end{equation}

Since the maximization is performed over subsets of the original domains, the discrete values form the lower bounds of their exact counterparts: \( \hat{\Delta}_v(s_v, s_o) \in [0, \Delta_v(s_v, s_o)], \quad \hat{\Delta}_o(s_v, s_o) \in [0, \Delta_o(s_v, s_o)] \). As a consequence, the global criterion also satisfies \( \hat{J}(s_v, s_o) \in [0, J(s_v, s_o)] \). 
This method yields a stochastic estimator of the exact criterion \( J(s_v, s_o) \), it reduces the computational cost of the evaluation process while preserving information and informative content.
The characteristics of the designed estimator $\hat{J}(s_v, s_o)$ suggest using evolutionary algorithms, which are well-suited for optimization problems with nondifferentiable and highly irregular cost functions.

\section{Solving the Game via Meta-Heuristics}
\label{sec:Meta-Heuristics}
When modeling complex game structures with nonlinear or non-convex functions, 
traditional solvers may not guarantee global convergence. Recent works 
\cite{le2021lucidgames,le2022algames} address this by employing an augmented 
Lagrangian formulation to solve constrained multi-player general-sum dynamic games, 
seeking points where the gradients of all augmented Lagrangian functions vanish. 
While offering near real-time performance and stable convergence, these solvers only 
guarantee local convergence satisfying first-order optimality conditions.

In light of these limitations, metaheuristic approaches emerge as a promising 
alternative. As emphasized in \cite{greiner2017game}, combining game theory with 
metaheuristics yields strong results for complex equilibrium problems, offering a 
balanced compromise between optimality and individual objectives 
\cite{vrahatis2020particle}. Among these, Particle Swarm Optimization (PSO) has 
proven particularly effective for multiplayer games with interconnected objectives. 
A concrete example is provided in \cite{youssefi2022swarm}, where game-theoretic 
PSO improves autonomous navigation in robotic search and rescue scenarios.
\subsection{Problem Reformulation in the PSO Context}
In this section, we propose an adaptation of the Particle Swarm Optimization (PSO) algorithm to search for a Nash equilibrium.
The goal of the PSO-based solver is to minimize \( \hat{J}(s_v, s_o) \) in Eq. \eqref{eq:JhateAppro}. 
Let a swarm of $P$ particles explore a search space of dimension $D$. Each particle $p \in \{1, \dots, P\}$ represents a candidate strategy profile of the form $(s_v, s_o)$ within the search space. Since we are seeking to compute a Nash equilibrium, each particle can be considered as a potential equilibrium candidate. We denote its associated strategic profile by:

\begin{equation}
\mathbf{X}_p = (s_v, s_o) \in S_v \times S_o.
\end{equation}

We decompose each strategy into two components, namely, finite-dimensional vectors $\mathcal{P}_p^{v,k}$ and $\mathcal{P}_p^{o,k}$ that parametrize the geometric paths of the autonomous vehicle (agent $v$) and the human driver $o$, respectively, and the corresponding velocity vectors $\mathcal{V}_p^{v,k}$ and $\mathcal{V}_p^{o,k}$ 
The explicit value of particle $p$ at generation $k$ is then defined as
$\mathbf{X}_p^k = \left( \mathcal{P}_p^{v,k}, \mathcal{V}_p^{v,k}, \mathcal{P}_p^{o,k}, \mathcal{V}_p^{o,k} \right)$. 
The objective of the PSO algorithm is to iteratively update 
these vectors according to PSO dynamics, in order to minimize 
the joint deviation measure $\widehat{J}$. For conciseness, 
we introduce the shorthand notation
\[
z_p^{v,k}=(\mathcal{P}_p^{v,k},\mathcal{V}_p^{v,k}),
\qquad
z_p^{o,k}=(\mathcal{P}_p^{o,k},\mathcal{V}_p^{o,k}),
\]
so that $\mathbf{X}_p^k = (z_p^{v,k}, z_p^{o,k})$ denotes 
the full strategy profile of particle $p$ at iteration $k$. 
The criterion to minimize is then:

\begin{equation}
\begin{aligned}
\widehat{J}(\mathbf{X}_p^k)
=&\;
\max\Bigg\{
\max_{\substack{
\bar{z}^{v,k}\in \widehat{S}_v\\
\bar{z}^{v,k}\neq z_p^{v,k}
}}
\Big[
Q_v(z_p^{v,k},z_p^{o,k})
-
Q_v(\bar{z}^{v,k},z_p^{o,k})
\Big],
0
\Bigg\}
\\
&+
\max\Bigg\{
\max_{\substack{
\bar{z}^{o,k}\in \widehat{S}_o\\
\bar{z}^{o,k}\neq z_p^{o,k}
}}
\Big[
Q_o(z_p^{v,k},z_p^{o,k})
-
Q_o(z_p^{v,k},\bar{z}^{o,k})
\Big],
0
\Bigg\}
\end{aligned}
\label{eq:Jhat}
\end{equation}

With $\widehat{S}_v$ and $\widehat{S}_o$ are finite-dimensional subsets obtained via random sampling from $S_v$ and $S_o$, respectively.

$\widehat{J}(\mathbf{X}_p^k)$ is an approximation of the continuous criterion $J(\mathbf{X}_p^k)$, which would result from computing the true supremum over $S_v$ and $S_o$, rather than through sampled maxima. Moreover, we note the important property that $\widehat{J}(\mathbf{X}_p^k) \in [0, J(\mathbf{X}_p^k)]$, which implies that the expectation $\mathbb{E}[\widehat{J}(\mathbf{X}_p^k)] \in [0, J(\mathbf{X}_p^k)]$.

\subsubsection*{\textbf{Definition of Personal Best and Global Best}}
\textcolor{white}{.}

We now outline the personal and global best solutions in the game-theoretic context:

\begin{itemize}
    \item \textbf{Personal Best} ($p_{\text{best},p}^k$): the best-known strategy profile found by particle $p$, i.e., the one that has yielded the smallest deviation from a Nash equilibrium:
\end{itemize}
\begin{equation}
\begin{aligned}
p_{\text{best},p}^k = (\mathcal{P}_p^{v,j}, \mathcal{V}_p^{v,j}, \mathcal{P}_p^{o,j}, \mathcal{V}_p^{o,j})
\end{aligned}
\end{equation}
Where $j = \arg\min_{1 \leq i \leq k} \widehat{J}(\mathcal{P}_p^{v,i}, \mathcal{V}_p^{v,i}, \mathcal{P}_p^{o,i}, \mathcal{V}_p^{o,i})$
\begin{itemize}
    \item \textbf{Global Best} ($g_{\text{best}}^k$): the best-known strategy profile across the entire swarm up to iteration $k$:
\end{itemize}
\begin{equation}
\begin{aligned}
g_{\text{best}}^k = (\mathcal{P}_p^{v,j}, \mathcal{V}_p^{v,j}, \mathcal{P}_p^{o,j}, \mathcal{V}_p^{o,j}),
\end{aligned}
\end{equation}
Where $(p,j) = \arg\min_{\substack{1 \leq i \leq k \\ 1 \leq q \leq P}} \widehat{J}(\mathcal{P}_q^{v,i}, \mathcal{V}_q^{v,i}, \mathcal{P}_q^{o,i}, \mathcal{V}_q^{o,i})$

Using these precise definitions of $p_{\text{best},p}^k$ and $g_{\text{best}}^k$, teach particle's velocity and position from iteration $k$ to $k + 1$ is governed by the following:

\begin{equation}
\begin{aligned}
\mathbf{V}_p^{k+1}
&=
\omega \mathbf{V}_p^k
+
c_1 r_1 \Delta\mathbf{p}_p^k
+
c_2 r_2 \Delta\mathbf{g}_p^k,
\\
\mathbf{X}_p^{k+1}
&=
\mathbf{X}_p^k
+
\mathbf{V}_p^{k+1}
\end{aligned}
\label{equ:update}
\end{equation}
where
\[
\Delta\mathbf{p}_p^k
=
\mathbf{p}_{best,p}^k-\mathbf{X}_p^k,
\qquad
\Delta\mathbf{g}_p^k
=
\mathbf{g}_{best}^k-\mathbf{X}_p^k.
\]

Here: $\omega$ is the inertia weight that governs the velocity's influence.
$c_1$, $c_2$ are cognitive and social learning coefficients \cite{gou2017novel}.
$r_1$, $r_2 \sim \mathcal{U}(0, 1)$ are stochastic, uniformly distributed, D-dimensional random vectors.


\section{Experimental Evaluation and Results}
\label{sec:XP}
\subsection{Experimental Protocol and Scenario}
The experimental validation was conducted on a closed test track reproducing a 
real-world unsignalized four-way intersection (lane width: 3.15\,m), located in 
Versailles-Satory, France. The setup involves two vehicles:

\begin{itemize}
    \item \textbf{Ego Vehicle (Autonomous, in red):} An experimental Renault Zoé 
    provided by the VEDECOM Institute, operating in autonomous mode, equipped with 
    LiDAR, Radar, and RTK-GNSS/INS localization. See Figure~\eqref{fig:Experimental 
    Automated}.
    
    \item \textbf{Opponent Vehicle (Human-Driven, in blue):} A standard vehicle 
    driven by a human, introducing realistic behavioral variability and 
    non-deterministic dynamics into the decision-making loop.
\end{itemize}

\begin{figure}[htp]
    \centering
     \includegraphics[width=0.35\textwidth]{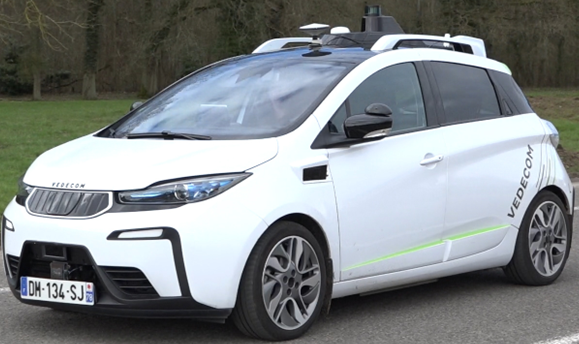}
    \caption{Experimental Automated Vehicle of VEDECOM}
    \label{fig:Experimental Automated}
    \end{figure}

\subsection{Scenario Configuration}
Both vehicles approach the intersection from opposing directions (Ego from the 
Northeast, Opp from the Southwest) to perform simultaneous left-turn maneuvers, 
creating a critical conflict zone that requires strategic coordination for collision-free passage.

\subsection{Strategic Adaptation to Human Behavior}
The framework was tested in scenarios requiring distinct strategic responses to 
the opponent's trajectory. Ego adapts its decisions based on Opp's predicted 
trajectory and movement through the conflict zone.

\subsubsection{\textbf{Simultaneous Crossing with Low-Speed Cooperative Adjustment}} 

\begin{figure}[htp]
\centering

\begin{minipage}[t]{0.49\linewidth}
  \includegraphics[width=\linewidth]{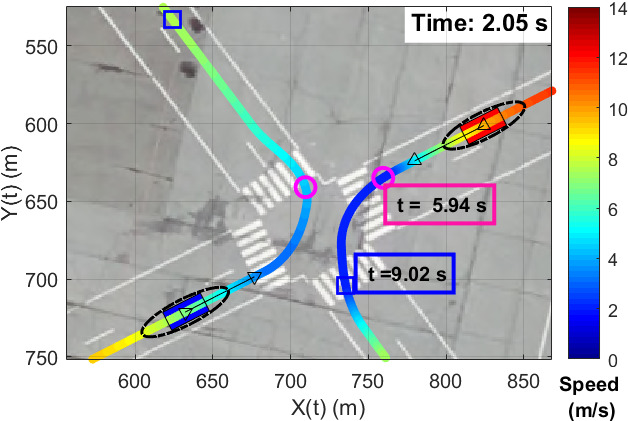}
  \centering \small (a) Top View -- Initial Positions  
\end{minipage}
\hfill
\begin{minipage}[t]{0.49\linewidth}
  \includegraphics[width=\linewidth]{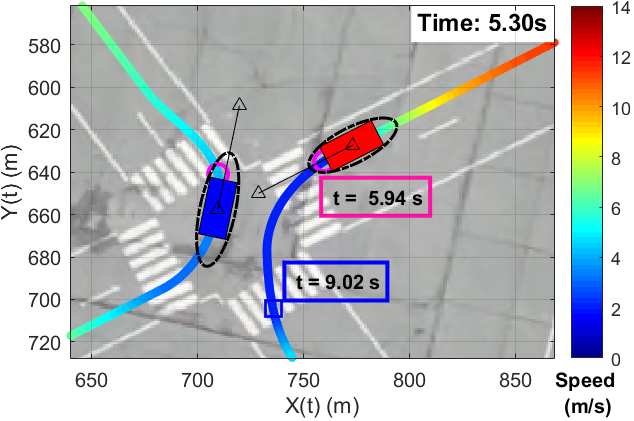}
  \centering \small (b) Top View -- Beginning of Maneuver 
\end{minipage}

\vspace{0.3em}

\begin{minipage}[t]{0.49\linewidth}
  \includegraphics[width=\linewidth]{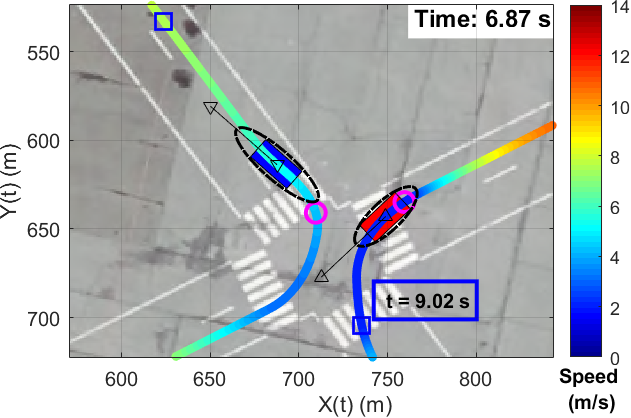}
  \centering \small (c) Top View -- Conflict Zone Interaction
\end{minipage}
\hfill
\begin{minipage}[t]{0.49\linewidth}
  \includegraphics[width=\linewidth]{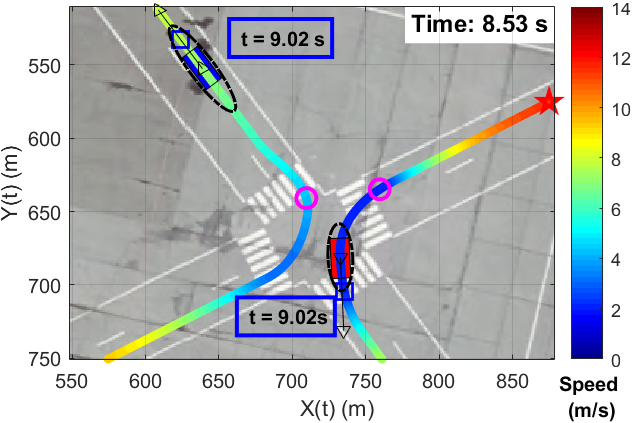}
  \centering \small (d) Top View -- End of Maneuver
\end{minipage}

\vspace{1em}

\begin{minipage}[t]{0.49\linewidth}
  \includegraphics[width=0.95\linewidth]{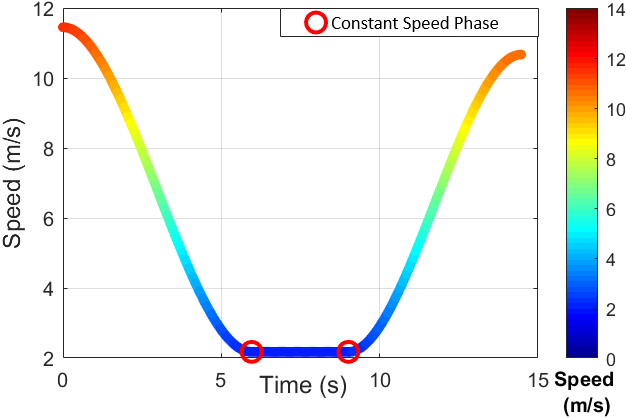}
  \centering \small (e) Ego vehicle speed over experiment time
\end{minipage}
\hfill
\begin{minipage}[t]{0.49\linewidth}
  \includegraphics[width=0.95\linewidth]{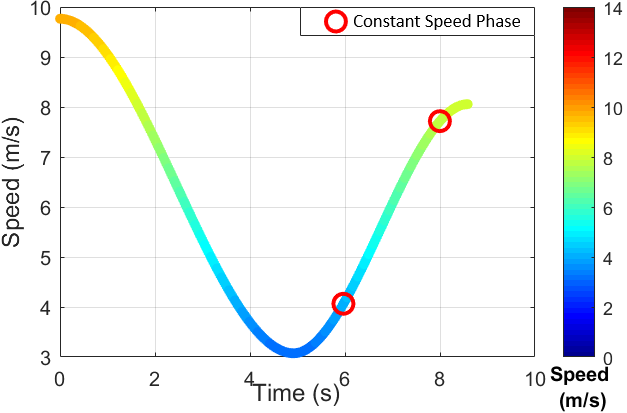}
  \centering \small (f) Opp vehicle speed over experiment time
\end{minipage}
\caption{Simultaneous Crossing with Low-Speed Maneuver}
\label{fig:interaction_case1}
\end{figure}

This scenario is illustrated in Figures~\ref{fig:interaction_case1}. It considers a simultaneous crossing maneuver where both agents negotiate 
access to a shared conflict zone. At $t \approx 2.0\,\text{s}$, Ego anticipates that 
Opp will clear the conflict area before its arrival, which is confirmed as Opp exits 
the shared zone by $t \approx 5.9\,\text{s}$. Ego then performs a controlled left-turn, 
crosses while Opp is already downstream ($t \approx 6.8\,\text{s}$), and completes 
the maneuver by $t \approx 9.0\,\text{s}$. The longitudinal profiles, illustrated in Figures~\ref{fig:interaction_case1}~(e--f), highlight the cooperative nature of this interaction: Opp 
follows a typical deceleration--acceleration pattern, while Ego maintains a steady 
low speed of $\approx 2\,\text{m/s}$ after an initial slowdown, avoiding unnecessary 
stops while preserving safety. The slight lateral deformation in Ego's trajectory 
reflects the safety-oriented cost terms, increasing lateral clearance without 
compromising comfort. Overall, this scenario illustrates anticipation-driven, 
non-aggressive coordination consistent with our game-theoretic framework.

\hspace*{-0.5cm}   

\subsubsection{\textbf{Early Crossing Uninterrupted Left Turn}}

\begin{figure}[H]
\centering

\begin{minipage}[t]{0.49\linewidth}
  \includegraphics[width=\linewidth]{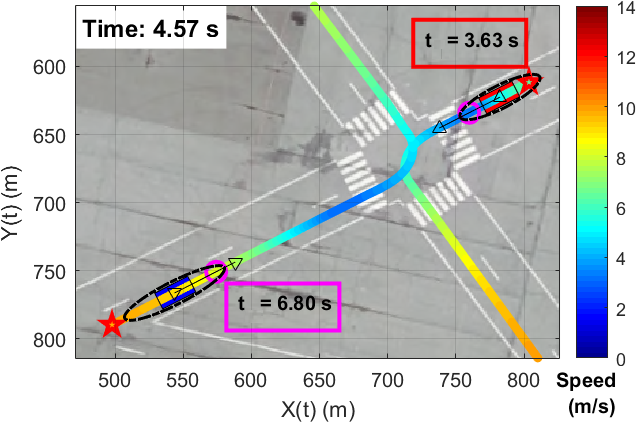}
  \centering \small (a) Top View -- Initial Positions  
\end{minipage}
\hfill
\begin{minipage}[t]{0.49\linewidth}
  \includegraphics[width=\linewidth]{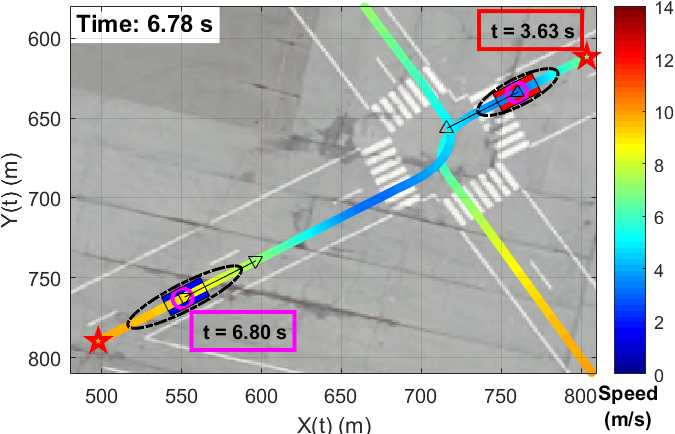}
  \centering \small (b) Top View -- Beginning of Maneuver 
\end{minipage}

\vspace{0.3em}

\begin{minipage}[t]{0.49\linewidth}
  \includegraphics[width=\linewidth]{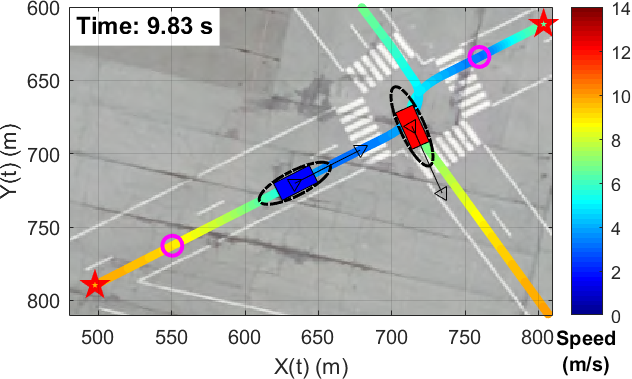}
  \centering \small (c) Top View -- Conflict Zone Interaction
\end{minipage}
\hfill
\begin{minipage}[t]{0.49\linewidth}
  \includegraphics[width=\linewidth]{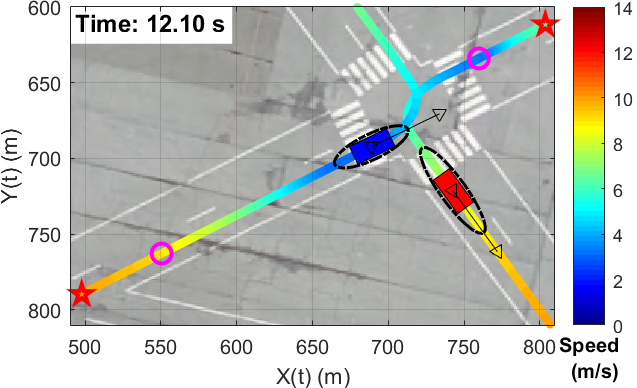}
  \centering \small (d) Top View -- End of Maneuver
\end{minipage}

\vspace{1em}


 \caption{Early Crossing Uninterrupted Left Turn by Ego}
\label{fig:interaction_case2}
 \end{figure}

In this scenario, illustrated in Figures~\ref{fig:interaction_case2}, Ego executes an early crossing while the human-driven Opp vehicle approaches from the Southwest. Initially, Opp is far from the intersection and does not create an immediate conflict. Ego begins decelerating to $3.5\,\text{m/s}$, adapting its speed to the intersection geometry.
Ego predicts its entry into the conflict zone around $t \approx 6.8\,\text{s}$, at which point Opp remains sufficiently upstream. Figures~\ref{fig:interaction_case2}~(c--d) show that Ego exits the intersection before Opp reaches the critical area.

\hspace*{-0.5cm}   

\subsection{Convergence Behavior and Runtime Assessment of the PSO-Based Solver}
\begin{figure}[H] 
  \centering
  \includegraphics[width=0.4\textwidth]{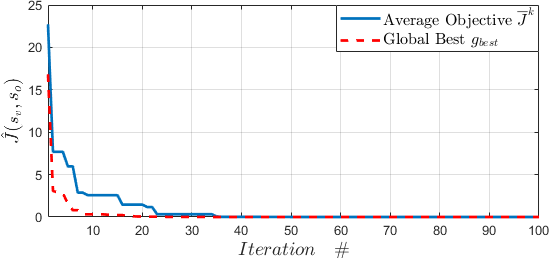}
  \caption{Convergence Profile – Use Case 2: Fast convergence with preserved exploration}
  \label{fig:VconstPSO}
\end{figure}

We assess the robustness, convergence behavior, and real-time feasibility of the 
proposed PSO-based solver across the two interaction scenarios studied above.
Figure~\eqref{fig:VconstPSO} illustrates convergence toward the Nash equilibrium, 
tracking both the global best particle cost $g_{\text{best}}$ and the swarm-wide 
average objective $\overline{J}_k = \frac{1}{P} \sum_{j=1}^{P} f(X^k_j)$. 
The solver exhibits fast initial convergence within the first 10 iterations, after which 
$g_{\text{best}}$ stabilizes near zero while the average objective $\overline{J}^k$ 
continues to decrease smoothly, reflecting preserved swarm exploration. Both metrics 
converge to a consistent Nash equilibrium, confirming solver robustness.

To further assess real-time feasibility, Table~\eqref{tab:PSO_solver_performance} 
reports the mean convergence time over 100 independent runs across configurations 
ranging from 50 to 160 strategies per agent, along with the $95\%$ confidence interval.

\begin{table}[H]
\centering
\renewcommand{\arraystretch}{1.2}
\caption{Computation Time and Equilibrium Accuracy of the PSO-Based Solver}
\label{tab:PSO_solver_performance}
\resizebox{\linewidth}{!}{%
\begin{tabular}{|c|c|c|c|c|}
\hline
\textbf{Scenario} & \textbf{Strategy Grid Size} & \textbf{Evaluations Used} & 
\textbf{Nash Cost} $\hat{J}(s_v, s_o)$ & \textbf{Time (s) ± Average (95\% CI)} \\
\hline
1 ( 100 Runs ) & $50 \times 50$   & $863 / 2500$     & $0.0000$ & $0.0195 \pm 0.0012$ \\
2 ( 100 Runs ) & $100 \times 100$ & $4603 / 10000$   & $0.0000$ & ${0.0203 \pm 0.0020}$ \\
3 ( 100 Runs ) & $100 \times 160$ & $5432 / 16000$   & $0.0000$ & $0.0312 \pm 0.0024$ \\
4 ( 100 Runs ) & $160 \times 160$ & $17200 / 25921$  & $0.0000$ & $0.0485 \pm 0.0024$ \\
\hline
\end{tabular}%
}
\vspace{0.5em}
\caption*{\small{\textit{Note:} The PSO solver consistently converges to a valid 
Nash equilibrium, with near real-time feasibility for strategy sets up to 
$160 \times 160$.}}
\end{table}

As expected, larger discretization grids yield longer convergence times. Notably, 
even for the largest tested configuration ($160 \times 160$), convergence is achieved 
in under $50$\,ms, confirming real-time suitability. Unlike fixed-grid approaches, 
the solver relies on random sampling, enabling stochastic estimation of the exact 
evaluation criteria within the game-theoretic framework.

\section{Conclusion}
This paper presented the experimental validation of a game-theoretic 
decision-making framework for autonomous vehicles in mixed traffic. Modeling 
interactions as a Generalized Nash Equilibrium Problem (GNEP) and solving it via 
a tailored Particle Swarm Optimization (PSO) algorithm, the approach generates 
interaction-aware strategies beyond traditional rule-based methods. Real-world 
experiments with an autonomous Renault Zoé at an unsignalized intersection confirmed 
the system's ability to negotiate simultaneous crossings and early merging scenarios 
under coupled safety constraints. The results validate both the GNEP formulation for 
capturing complex traffic dependencies and the solver's real-time efficiency 
($<50$\,ms convergence).

\textbf{Extension to Multi-Agent Scenarios
}: The current framework is formulated as a two-player game, covering the
most critical interaction dyad in intersection crossing. Extending it
to $N > 2$ players raises two challenges: the PSO search space grows
combinatorially to $\sum_{i=1}^{N} d_i$ dimensions, and the shared
constraint set $h(s_1, \ldots, s_N) \leq 0$ must encode pairwise
collision avoidance between all agents, potentially rendering the joint
feasible set non-convex. Both aspects will be investigated in future
work through simulation in multi-vehicle interaction.
scenarios.
 \bibliographystyle{IEEEtran}
 \bibliography{these}   

\begin{thebibliography}{10}
\providecommand{\url}[1]{#1}
\csname url@samestyle\endcsname
\providecommand{\newblock}{\relax}
\providecommand{\bibinfo}[2]{#2}
\providecommand{\BIBentrySTDinterwordspacing}{\spaceskip=0pt\relax}
\providecommand{\BIBentryALTinterwordstretchfactor}{4}
\providecommand{\BIBentryALTinterwordspacing}{\spaceskip=\fontdimen2\font plus
\BIBentryALTinterwordstretchfactor\fontdimen3\font minus
  \fontdimen4\font\relax}
\providecommand{\BIBforeignlanguage}[2]{{%
\expandafter\ifx\csname l@#1\endcsname\relax
\typeout{** WARNING: IEEEtran.bst: No hyphenation pattern has been}%
\typeout{** loaded for the language `#1'. Using the pattern for}%
\typeout{** the default language instead.}%
\else
\language=\csname l@#1\endcsname
\fi
#2}}
\providecommand{\BIBdecl}{\relax}
\BIBdecl

\bibitem{mahajan2021intention}
A.~Mahajan, T.~Kumano, and Y.~Yasui, ``Intention estimation and controllable
  behaviour models for traffic merges,'' \emph{Journal of Control, Measurement,
  and System Integration}, 2021.

\bibitem{qin2024game}
Z.~Qin, A.~Ji, Z.~Sun, G.~Wu, P.~Hao, and X.~Liao, ``Game theoretic application
  to intersection management: A literature review,'' \emph{IEEE Transactions on
  Intelligent Vehicles}, 2024.

\bibitem{zhang2021safe}
Z.~Zhang and J.~F. Fisac, ``Safe occlusion-aware autonomous driving via
  game-theoretic active perception,'' in \emph{17th Robotics: Science and
  Systems, RSS 2021}.\hskip 1em plus 0.5em minus 0.4em\relax MIT Press
  Journals, 2021.

\bibitem{bacharach1989zero}
M.~Bacharach, ``Zero-sum games,'' \emph{Game theory}, pp. 253--257, 1989.

\bibitem{8643396}
A.~Liniger and J.~Lygeros, ``A noncooperative game approach to autonomous
  racing,'' \emph{IEEE Transactions on Control Systems Technology}, vol.~28,
  no.~3, pp. 884--897, 2020.

\bibitem{nash1950equilibrium}
J.~F. Nash~Jr, ``Equilibrium points in n-person games,'' \emph{Proceedings of
  the national academy of sciences}, vol.~36, no.~1, pp. 48--49, 1950.

\bibitem{lei2022stochastic}
J.~Lei and U.~V. Shanbhag, ``Stochastic nash equilibrium problems: Models,
  analysis, and algorithms,'' \emph{IEEE Control Systems Magazine}, vol.~42,
  no.~4, pp. 103--124, 2022.

\bibitem{facchinei2010generalized}
F.~Facchinei and C.~Kanzow, ``Generalized nash equilibrium problems,''
  \emph{Annals of Operations Research}, vol. 175, no.~1, 2010.

\bibitem{naidja2025toward}
N.~Naidja, S.~Font, M.~Revilloud, and G.~Sandou, ``Toward a holistic
  multi-criteria trajectory evaluation framework for autonomous driving in
  mixed traffic environment,'' in \emph{2025 8th International Conference on
  Intelligent Robotics and Control Engineering (IRCE)}.\hskip 1em plus 0.5em
  minus 0.4em\relax IEEE, 2025, pp. 1--10.

\bibitem{askari2019behavioral}
G.~Askari, M.~E. Gordji, and C.~Park, ``The behavioral model and game theory,''
  \emph{Palgrave Communications}, vol.~5, no.~1, 2019.

\bibitem{le2021lucidgames}
S.~Le~Cleac’h, M.~Schwager, and Z.~Manchester, ``Lucidgames: Online unscented
  inverse dynamic games for adaptive trajectory prediction and planning,''
  \emph{IEEE Robotics and Automation Letters}, vol.~6, no.~3, pp. 5485--5492,
  2021.

\bibitem{le2022algames}
------, ``Algames: a fast augmented lagrangian solver for constrained dynamic
  games,'' \emph{Autonomous Robots}, vol.~46, no.~1, pp. 201--215, 2022.

\bibitem{greiner2017game}
D.~Greiner, J.~Periaux, J.~M. Emperador, B.~Galv{\'a}n, and G.~Winter, ``Game
  theory based evolutionary algorithms: a review with nash applications in
  structural engineering optimization problems,'' \emph{Archives of
  Computational Methods in Engineering}, vol.~24, pp. 703--750, 2017.

\bibitem{vrahatis2020particle}
M.~N. Vrahatis, P.~Kontogiorgos, and G.~P. Papavassilopoulos, ``Particle swarm
  optimization for computing nash and stackelberg equilibria in energy
  markets,'' in \emph{SN Operations Research Forum}, vol.~1, no.~3.\hskip 1em
  plus 0.5em minus 0.4em\relax Springer, 2020, p.~20.

\bibitem{youssefi2022swarm}
K.~A.-R. Youssefi, M.~Rouhani, H.~R. Mashhadi, and W.~Elmenreich, ``A swarm
  intelligence-based robotic search algorithm integrated with game theory,''
  \emph{Applied Soft Computing}, vol. 122, p. 108873, 2022.

\bibitem{gou2017novel}
J.~Gou, Y.-X. Lei, W.-P. Guo, C.~Wang, Y.-Q. Cai, and W.~Luo, ``A novel
  improved particle swarm optimization algorithm based on individual difference
  evolution,'' \emph{Applied Soft Computing}, vol.~57, pp. 468--481, 2017.

\end{thebibliography}
\end{document}